\documentclass[10pt,twocolumn,letterpaper]{article}

\usepackage{wacv}              

\usepackage{graphicx}
\usepackage{amsmath}
\usepackage{amssymb}
\usepackage{booktabs}
\usepackage{adjustbox} 

\usepackage[accsupp]{axessibility}  

\usepackage[pagebackref,breaklinks,colorlinks]{hyperref}

\usepackage[capitalize]{cleveref}
\crefname{section}{Sec.}{Secs.}
\Crefname{section}{Section}{Sections}
\Crefname{table}{Table}{Tables}
\crefname{table}{Tab.}{Tabs.}

\def\wacvPaperID{3} 
\def\confName{WACV}
\def\confYear{2025}

\begin{document}

\title{MedGrad E-CLIP: Enhancing Trust and Transparency in AI-Driven Skin Lesion Diagnosis}

\author{Sadia Kamal\\
University of Maryland Baltimore County\\
{\tt\small sadia1402@umbc.edu}
\and
Tim Oates\\
University of Maryland Baltimore County\\
{\tt\small oates@umbc.edu}
}

\maketitle

\begin{abstract}
As deep learning models gain attraction in medical data, ensuring transparent and trustworthy decision-making is essential. In skin cancer diagnosis, while advancements in lesion detection and classification have improved accuracy, the black-box nature of these methods poses challenges in understanding their decision processes, leading to trust issues among physicians. This study leverages the CLIP (Contrastive Language-Image Pretraining) model, trained on different skin lesion datasets, to capture meaningful relationships between visual features and diagnostic criteria terms. To further enhance transparency, we propose a method called MedGrad E-CLIP, which builds on gradient-based E-CLIP by incorporating a weighted entropy mechanism designed for complex medical imaging like skin lesions. This approach highlights critical image regions linked to specific diagnostic descriptions. The developed integrated pipeline not only classifies skin lesions by matching corresponding descriptions but also adds an essential layer of explainability developed especially for medical data. By visually explaining how different features in an image relates to diagnostic criteria, this approach demonstrates the potential of advanced vision-language models in medical image analysis, ultimately improving transparency, robustness, and trust in AI-driven diagnostic systems.


\end{abstract}


\section{Introduction}
\label{sec:intro}

Cancer is characterized by the uncontrolled growth of body cells and is a major global health concern. Among its various forms, skin cancer is the most common, primarily affecting areas of the body frequently exposed to the sun. The primary cause of skin cancer is excessive exposure to ultraviolet (UV) radiation, which can lead to life-threatening conditions in as little as six weeks.
Early identification of skin diseases is critical as it can significantly improve outcomes and reduce healthcare costs.

Various methods have been developed to detect and differentiate skin lesion types \cite{esteva2017dermatologist} \cite{adegun2021deep} \cite{gessert2020skin} \cite{ali2021enhanced} \cite{salopek2001differentiation}.
Melanomas, the most serious form of skin cancer, exhibit a range of characteristics, such as the presence or absence of pigmentation and diagnostic features like the whitish veil.
Clinicians have established different guidelines such as the ABCDE rule—Asymmetry, Border irregularity, Color variation, Diameter, and Evolution—to track changes in lesions \cite{nachbar1994abcd}.  However, image resolution variations can complicate diameter assessment, and these features alone may not ensure accurate diagnosis of different types of melanomas. Consequently, the Menzies method was developed as a simplified dermoscopy technique for melanoma diagnosis,by focusing on key "negative" and "positive" dermoscopic features \cite{singh2014detection}.
Despite its improved accuracy over the ABCDE rule, the Menzies method has high sensitivity \cite{masood2013computer} \cite{carrera2016validity}, which can lead to false-positives, 
especially when used by less experienced clinicians.
To overcome the limitations of the Menzies method, the 7-point checklist was introduced \cite{walter2013using}.
However, this method also presents challenges for non-experts, as accurate diagnosis without specialized tools is difficult.

The complexity of diagnosing skin lesions highlights the need for manual evaluation by clinicians. Nonetheless, automated techniques using deep neural networks offer promising solutions by improving the precision and reliability of skin lesion detection and classification \cite{lopez2017skin} \cite{zhang2019attention}. Despite their potential, these methods are often perceived as 'black boxes,' which makes it challenging for clinicians to trust their outputs. While some studies have focused on enhancing the explainability of medical data to build transparency and trust \cite{ge2017skin} \cite{nigar2022deep}, they have not addressed the importance of highlighting specific regions in relation to their corresponding textual descriptions, which would further enhance explainability and interpretability. Furthermore, no existing classification method fully integrates all diagnostic techniques.

To address this research gap, we developed a pipeline aimed at assisting clinicians by training CLIP \cite{radford2021learning} on dermoscopic skin lesions images and their descriptions, incorporating features from all diagnostic techniques. Additionally, we introduced MedGrad E-CLIP, a novel explainability approach for skin lesion analysis that enhances the gradient-based CLIP \cite{zhao2024gradient} method by emphasizing fine-grained features crucial for accurate diagnosis. This method highlights subtle changes in lesion images by computing entropy, identifying specific regions relevant to each diagnosis and aligning them with corresponding textual descriptions, thus bridging visual data with diagnostic terms. This enhanced transparency promotes trust among clinicians, enabling them to understand and verify the AI's decision-making process.

\section{Related Works}

\subsection{CLIP in Medical Data}

CLIP has gained attraction in medical domains with models like Med-CLIP \cite{wang2022medclip} for semantic matching, eCLIP \cite{kumar2024improving} using radiologist eye-gaze data, Mammo-CLIP \cite{chen2024mammo} for multi-view mammograms, and ConVIRT \cite{zhang2022contrastive} for unsupervised pretraining with image-text pairs. MITER (Medical Image–TExt joint adaptive pRetraining) \cite{shu2024miter} combines multi-level contrastive learning, pathCLIP \cite{he2024pathclip} enhances gene identification via image-text embeddings , CLIPath \cite{lai2023clipath} fine-tunes CLIP with residual connections, and PubMedCLIP \cite{eslami2023pubmedclip} excels in medical VQA tasks. Despite these advancements, several research gaps remain in using CLIP models for medical tasks, particularly concerning generalizability across diverse medical domains and explainability.

\subsection{Explainability in Medical Data}

The growing use of deep learning in medical image detection, classification, and segmentation has raised concerns about the black-box nature of these models, making trust and transparency crucial for clinician acceptance.


Numerous studies have focused on enhancing the explainability of models applied to medical data \cite{munn2022explainable}. In the realm of image-based explanations, the primary objective is to identify the specific parts of an image (pixels or segments) that most significantly influence a model's prediction. Prominent techniques include gradient-based methods for convolutional neural networks (CNNs), such as Guided Backpropagation, CAM, Grad-CAM \cite{selvaraju2017grad}, GradCAM++ \cite{chattopadhay2018grad}, Guided GradCAM \cite{selvaraju2016grad}, SmoothGrad \cite{smilkov2017smoothgrad} and DeepLIFT \cite{li2021deep}. These were developed to further enhance the interpretability of model predictions by offering more refined visual explanations that highlight the regions of the input most responsible for the model's decisions in medical data.

In addition to gradient-based methods, other approaches like SHAP (SHapley Additive exPlanations) \cite{lundberg2017unified}, LIME (Local Interpretable Model-agnostic Explanations) \cite{ribeiro2016should} and Layer-wise Relevance Propagation (LRP) \cite{bach2015pixel} have been developed to provide more generalizable explanations across different types of data. These techniques offer insights into the contribution of individual features to the model's predictions. However, some medical data often rely on multiple data sources such as images, EHR and clinical notes. Grad E-Clip \cite{zhao2024gradient}, an emerging gradient based method which provides comprehensive explanations across different modalities by highlighting important areas in the images. It works well for general images but struggles with complex medical images, where small, important details often go unnoticed. This highlights the need for more precise explainability methods for reliable medical image analysis.

\section{Method Overview}
\subsection{Dataset}

Based on our work requirement which uses images with specific dermoscopic structure criteria, we used the PH² and Derm7pt datasets. The PH² \cite{mendoncca2013ph} image database contains a total of around 200 dermoscopic images of melanocytic lesions, including common nevi, atypical nevi, and melanomas. These includes clinical and histological diagnoses and the identification of several dermoscopic structure criteria.

Similarly, Derm7pt \cite{kawahara2019skin} is a dermoscopic image dataset that contains over 2000 clinical and dermoscopy images along with corresponding structured metadata tailored for training and evaluating computer aided diagnosis (CAD). This dataset includes the 7-point checklist for assessing the malignancy of skin lesions, making it a valuable resource for our study.

\subsection{Data Prepration}
The dataset is organized with each row as a unique image-text pair, removing duplicates to prevent overfitting. For text preprocessing, special characters and unnecessary punctuation are removed. While, images are resized to 224x224 pixels to meet the input requirements of the image encoder. To increase the number of image-text pairs, augmentations are applied: images are augmented through flipping and rotating, while text descriptions are reordered to create variations for the same image. These augmented text descriptions are then tokenized to create a format compatible with the text encoder, splitting the text into tokens (words). This careful pairing and preprocessing of images and text is crucial, as CLIP relies on learning the relationships between image-text pairs to function effectively


\subsection{Contrastive Learning Image Pretained - CLIP}
Contrastive Language-Image Pre-training (CLIP) consists of two key components: an image encoder and a text encoder, both of which are jointly trained to extract feature embeddings from images and text into a shared representation space. In this study, a pre-trained model with a vision transformer (ViT) is used as an image encoder, while a transformer-based encoder is used for text. Given an image-text pair $(I, T)$, the matching score between their extracted image features $f_I \in \mathbb{R}^D$ and text features $f_T \in \mathbb{R}^D$ is:
\begin{equation}
S(f_I, f_T) = \cos(f_I, f_T) = \frac{f_I f_T^T}{\|f_I\| \|f_T\|}
\end{equation}

\begin{figure}[h]  
    \centering
    \includegraphics[width=0.4\textwidth]{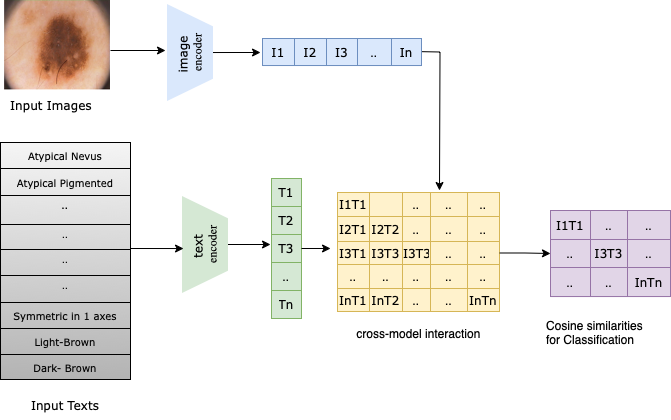}    \caption{\textbf{CLIP Overview for Custom Dataset:}
    We encode skin lesion images and their descriptions to generate image and text embeddings. These are combined in a cross-modal interaction module, calculating cosine similarities to assess alignment between lesions and diagnoses, ensuring accurate classification.}
    \label{fig:CLIP_overview}
\end{figure}

As shown in Figure \ref{fig:CLIP_overview}, CLIP
was trained on colored dermoscopic images paired with their structure criteria, which served as descriptive annotations. These trained weights were then employed for the classification of new image-text pairs.

Formally, let $f_x$ represent the image features extracted by CLIP's image encoder for lesion image $x$. The text features, which include descriptive criteria such as "melanoma", "symmetry",etc., are extracted using CLIP's text encoder, resulting in a set of $w_i$ features $W = \{w_i\}_{i=1}^K$, where $K$ represents the number of classes, such as Melanoma, Atypical Nevus, Common Nevus, etc., along with their descriptions. The probability of predicting class $i$ (e.g., melanoma) given input image $x$ is computed as:
\vspace{-2mm}

\begin{equation}
p(y = i | x) = \frac{\exp(\cos(w_i, f_x) / \tau)}{\sum_{j=1}^K \exp(\cos(w_j, f_x) / \tau)},
\end{equation}

where $\cos(\cdot, \cdot)$ denotes the cosine similarity between two vectors, and $\tau$ is a scaling factor learned by CLIP \cite{wu2023eventclip}. During training, the model learns to maximize the cosine similarity between the image features $f_x$ and the correct text features $w_i$ for the true class. Simultaneously, it minimizes the cosine similarity between $f_x$ and text features $w_j$ for all incorrect classes j $\neq$ i. This aligns the image and text embeddings in the feature space, enhancing the model's ability to accurately match images with their corresponding diagnoses.

\subsection{MedGrad E-CLIP}

This study presents an explainability method, specifically designed for analyzing medical imagery, such as skin lesions. While Grad E-CLIP highlights regions with high gradient values, it mainly relies on 
the loosened similarity function which reduces the sparsity issue of softmax attention by allowing more distributed and significant spatial importance. However, it still tends to bias attention towards dominant regions, often overlooking subtle but diagnostically significant details. This limitation can be particularly problematic when using the CLIP model, trained on skin lesion dataset, as it may miss important diagnostic information, thus reducing interpretability.

To address this, we introduce a weighted entropy mechanism, replacing Grad E-CLIP's spatial importance. Our approach moves beyond traditional spatial weighting by using entropy to directly measure the variability or complexity within regions, allowing the model to distribute focus across areas with subtle and rich information.  In this method, entropy is calculated locally to measure the variability and uncertainty of pixel intensity values within a defined neighborhood. It quantifies the information content by evaluating the distribution of gray-level values, capturing the complexity and richness of local image patterns. The entropy weight (\(w_e\)) is derived from the entropy values of each region, where higher entropy values correspond to regions with greater pixel-level complexity and information density. These weights allow the model to focus more on subtle patterns that are often diagnostically significant but overlooked by gradient-based attention mechanisms. As shown in Eq.~(\ref{eq:3}):

\begin{equation}
H_i = \text{ReLU}\left(\sum_{c} w_c v_i w_e \right)
\label{eq:3}
\end{equation}

\noindent where $w_c$ is the channel importance , $v_i$ represents the pixel values at spatial location \textit{i} and $w_e$ represents the computed entropy weights.

This results in a more balanced representation, enhancing the model's robustness and interpretability, especially in medical imaging where significant details and changes are crucial for accurate diagnosis. By introducing entropy weighting, the model shifts its focus towards regions with higher complexity and variability rather than merely the most activated areas, enabling a more comprehensive analysis of subtle diagnostic features and ultimately improving the explainability and effectiveness of medical image analysis.

\section{Our Approach}

\begin{figure}[h]  
    \centering
    \includegraphics[width=0.45\textwidth]{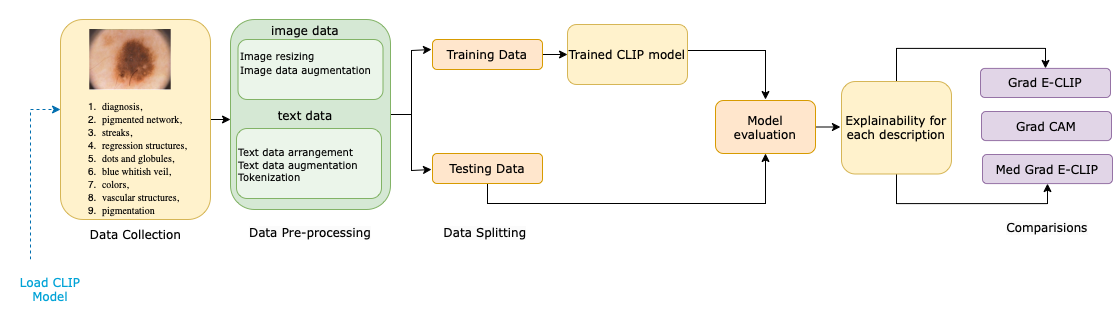}  
    \caption{Proposed pipeline}
    \label{fig:pipeline}
\end{figure}

In this study, we have developed a fully connected pipeline for classifying and differentiating various skin lesions, leveraging a dual-modality approach that integrates image and textual data. As shown in Figure \ref{fig:pipeline}, data were collected from two different databases, including images and their corresponding text descriptions. The collected images and text were then pre-processed, with images resizing, text organization, and data augmentation applied to both modalities. The dataset comprised of 17 distinct skin lesion types as classes. After pre-processing, 75\% of the dataset was allocated for training and the remaining 25\% for testing. The data was trained for 30 epochs with a batch size of 64, using the Adam optimizer with a learning rate of 1e-5. These settings achieved the optimal results during hyperparameter tuning. The loss use was the mean of image and text cross-entropy.

After training, the updated model weights were used to evaluate the model's performance on the test dataset. To enhance the interpretability of the newly trained CLIP model, explainability techniques were applied, offering both visual and textual insights into the decision-making processes of the model. These explainability approaches were subsequently compared to evaluate their effectiveness.


Our framework extends beyond just applying existing explainability techniques like Grad CAM, and Grad E-CLIP, to a pre-trained CLIP model by developing a comprehensive classification framework that effectively integrates image and text pairs. A critical aspect of our methodology is the development of MedGrad E-CLIP to incorporate computed entropy weights, replacing traditional spatial importance metrics. Our modified approach computes a local entropy within the region covered by the disk around each pixel that emphasizes minute changes within the lesion images. This adjustment allows for a more detailed understanding of the model’s decision-making process.
By enhancing both the accuracy and interpretability of classifications, our methodology emphasizes the critical role of transparency in AI-driven diagnostics, making a significant contribution especially to AI-driven medical diagnostics.

\section{Experiments}
Our experiments were performed out on Google Colab, utilizing a TESLA T4 GPU. We conducted these experiments using the ViT-B/16 architecture, which is based on a transformer model with a 16x16 patch size. The experiments were divided into two main parts i.e. performance of CLIP and performance of Explainability on CLIP.

\subsection{Performance Evaluation of the CLIP Model}
In this part we evaluate the performance of the CLIP model trained on custom skin lesion dataset. The training and testing performances are evaluated allowing for a direct comparison of their performance. The evaluation metrics for the training data are presented in Table \ref{table:training_metrics}, and Table \ref{table:test_metrics} compares test data performance before and after training on skin-lesion data.
\vspace{-2mm}
\begin{table}[h!]
\scriptsize  
\centering
\begin{tabular}{|c|c|}
\hline
\textbf{Evaluation Metrics} & \textbf{Value} \\
\hline
Accuracy & 81.80\% \\
\hline
Loss & 0.4771 \\
\hline
Precision & 0.8195 \\
\hline
Recall & 0.818 \\
\hline
F1-score & 0.8179 \\
\hline
Sensitivity & 0.818 \\
\hline
Specificity & 0.9971 \\
\hline
\end{tabular}
\caption{Model metrics on training data}
\label{table:training_metrics}
\end{table}
\vspace{-2mm}
\begin{table}[h!]
\scriptsize  
\centering
\begin{tabular}{|c|c|c|}
\hline
& \textbf{Before Training} & \textbf{After Training} \\
\hline
Number of test samples & 1215 & 1215 \\
\hline
Batch size & 64 & 64 \\
\hline
Accuracy & 2.06\% & 80.08\% \\
\hline
F1-Score & 0.0153 & 0.8011 \\
\hline
Average Loss & 4.1579 & 0.4954 \\
\hline
Average CLIP Score & 0.3081 & 0.9655 \\
\hline
\end{tabular}
\caption{Model metrics on test data before and after training on custom data}
\label{table:test_metrics}
\end{table}
\vspace{-3mm}

Accuracy, being the most commonly used metric, evaluates the overall performance of deep learning models. In addition to accuracy, other evaluation metrics such as Sensitivity, Specificity, Precision, and F1-score are also assessed for the CLIP model. These provide a more comprehensive evaluation of the model's performance by offering insights into its ability to correctly identify true positives, avoid false negatives, and maintain a balance between precision and recall. Furthermore, the CLIP Score (\(S_{\text{CLIP}}\)) is calculated as the cosine similarity between the image and text embeddings.
\vspace{-1mm}
\begin{equation}
S_{\text{CLIP}} = \frac{f_{\text{img}}(I) \cdot f_{\text{text}}(T)}{\|f_{\text{img}}(I)\| \|f_{\text{text}}(T)\|}
\end{equation}

\begin{figure}[htbp]
    \centering
    \includegraphics[width=0.5\linewidth]{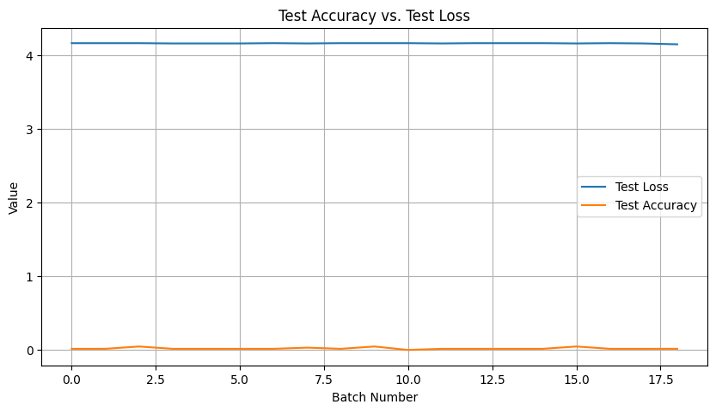}
    \caption{Test accuracy and loss on pre-trained model}
    \label{fig:pre-loss}
\end{figure}

\begin{figure}[htbp]
    \centering
    \includegraphics[width=0.5\linewidth]{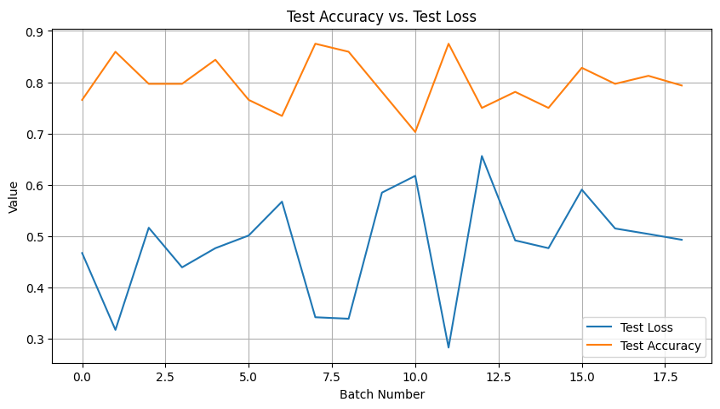}
    \caption{Test accuracy and loss on fine-tuned model}
    \label{fig:fine-loss}
\end{figure}

These metrics indicate the effectiveness of the learning algorithm, as the training curves reach a point of stability. In contrast, Figure \ref{fig:pre-loss} and \ref{fig:fine-loss} shows the learning curves for test accuracy and loss before and after training the model on the custom dataset, respectively. It is clear that the performance of the CLIP model improves significantly on trained model, leading to enhanced classification performance.

\subsection{Performance Evaluation of the MedGrad E-CLIP}
The second part of the experiments focused on evaluating the explainability of the CLIP models both before and after training. Gradient-based explainability methods, such as Grad-CAM, Grad E-CLIP and our proposed MedGrad E-CLIP explainability method, were employed to analyze the image-text pair understanding of model’s decision-making process on the skin lesion dataset. The results indicate that trained CLIP model on our custom data not only improved the model’s accuracy on the skin lesion dataset but also influenced the explainability of the model's outputs. In this study weighted entropy ensures that the model’s attention is distributed across not only the most prominent features but also those regions that provide critical, subtle information, regardless of texture and detail variations within the images.

\begin{figure*}[h]  
    \centering
    \tiny Atypical Nevus
    \adjustbox{valign=t,center}{\includegraphics[width=0.10\linewidth]{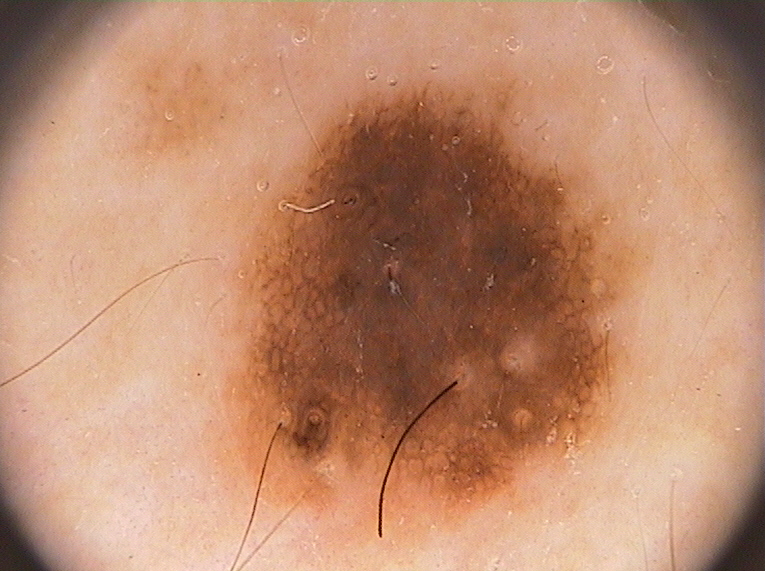}}
    \includegraphics[width=0.80\linewidth]{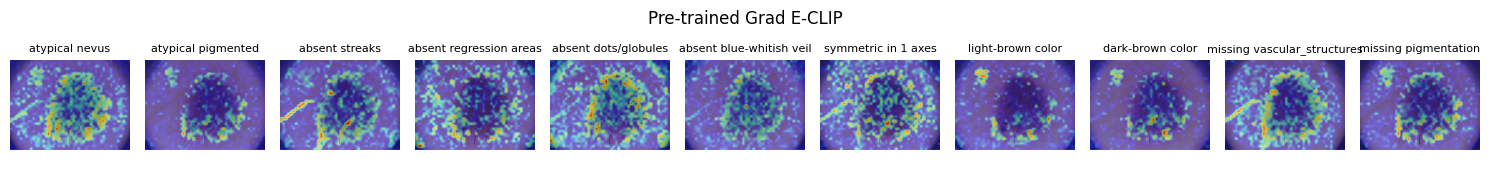}
    \includegraphics[width=0.80\linewidth]{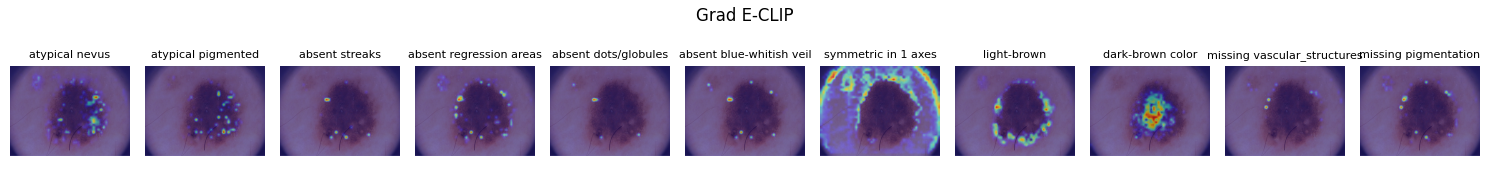}
    \includegraphics[width=0.80\linewidth]{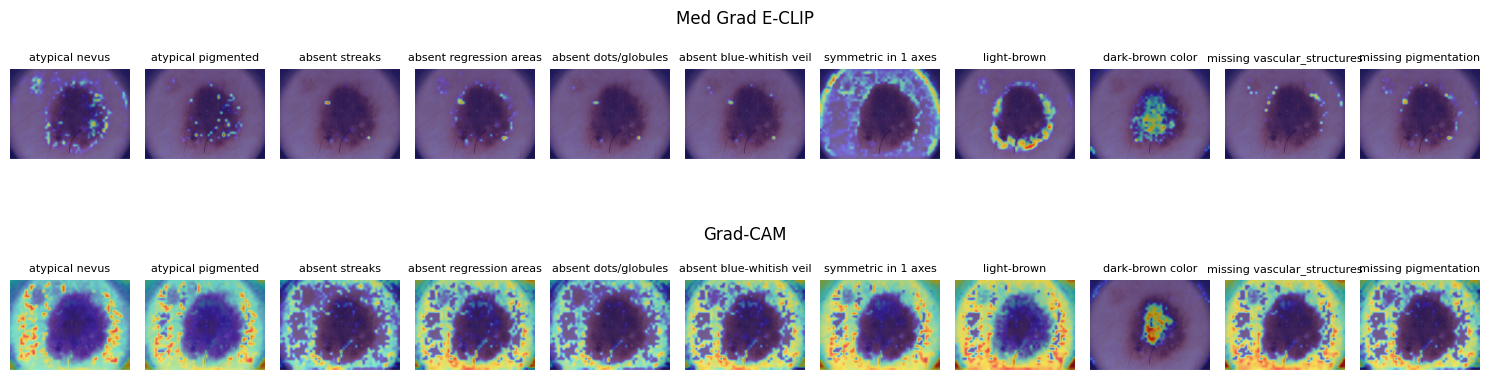}
    \caption{\small Comparative visualization of explainability methods—Original Atypical Nevus, Pre-trained Grad E-CLIP, Trained Grad E-CLIP, MedGrad E-CLIP, and Grad-CAM on \textbf{Atypical Nevus}}
    \label{fig:atypical_finetuned}
\end{figure*}

\begin{figure*}[h]  
    \centering
    \tiny Melanoma
    \adjustbox{valign=t,center}
    {\includegraphics[width=0.10\linewidth]{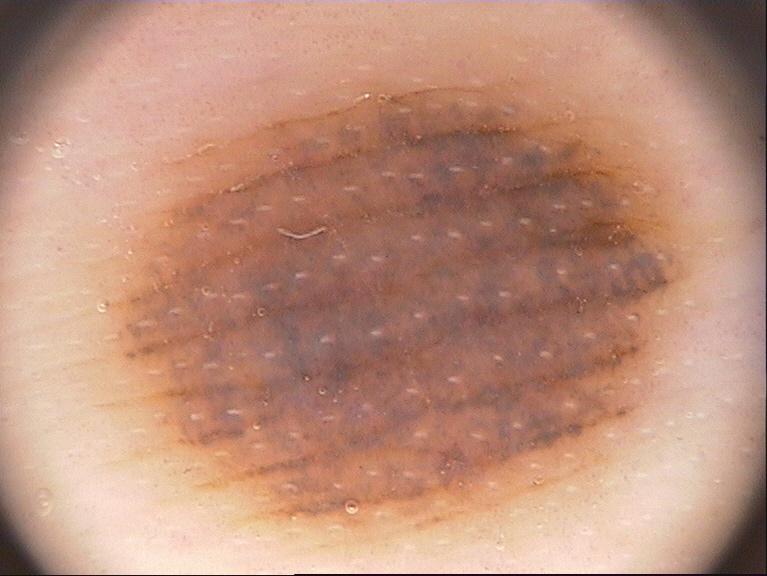}}
    \includegraphics[width=0.80\linewidth]{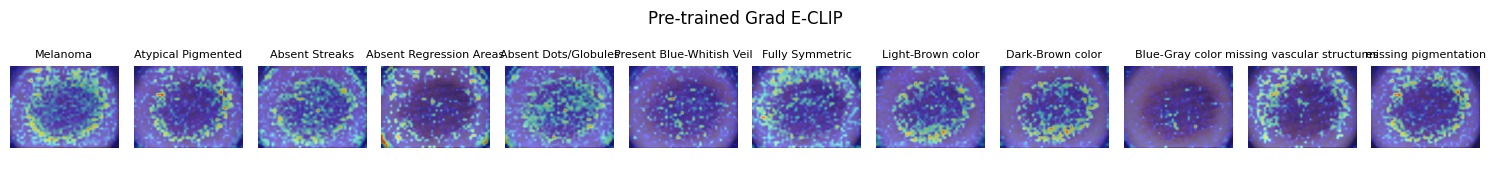}
    \includegraphics[width=0.80\linewidth]{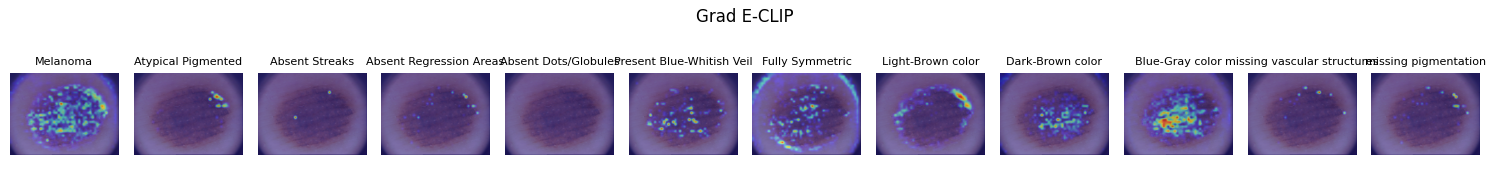}
    \includegraphics[width=0.80\linewidth]{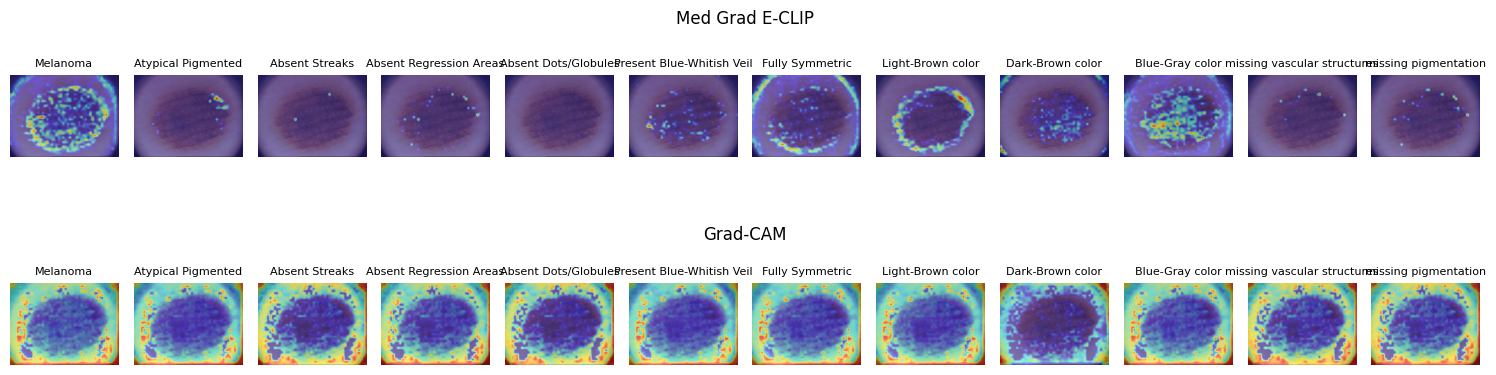}
    \caption{\small Comparative visualization of explainability methods—Original Melanoma, Pre-trained Grad E-CLIP, Trained Grad E-CLIP, MedGrad E-CLIP, and Grad-CAM on \textbf{Melanoma}}
    \label{fig:melanoma_finetuned}
\end{figure*}

Figures \ref{fig:atypical_finetuned}, and \ref{fig:melanoma_finetuned} shows the explainability results after training on custom skin lesion data. Figure \ref{fig:atypical_finetuned} is the explainability achieved on Atypical Nevus conditions while Figure \ref{fig:melanoma_finetuned} shows the explainability of Melanoma conditions. Each column in these figures corresponds to a specific skin lesion type, annotated with characteristics such as "typical pigmented" or "absent streaks". The first row is the original image with skin lesion conditions. The second row of these figures presents the Grad E-CLIP results from the pre-trained CLIP model. Third row shows the Grad E-CLIP visualizations on custom trained model. The fourth row shows our proposed MedGrad E-CLIP explainability highlighting areas of the image that contribute to the model's predictions and the last row shows the Grad-CAM visualizations.For the CLIP model, Grad-CAM was evaluated based on the cosine similarity of the image-text pair, using the gradients calculated with respect to the patch tokens from the ViT layers.

These heatmaps shows that our MedGrad E-CLIP method provides superior explainability by capturing subtle changes along with their relation to each input text compared to Grad-CAM and original Grad E-Clip, where some information were getting lost.






As discussed in \cite{zhao2024gradient}, the Grad E-CLIP model excels at identifying common perceptual attributes such as color, but it struggles with physical attributes like shape and material, and is less effective at grounding objects with comparative attributes, like size and positional relationships. Our proposed method addresses these limitations of the Grad E-CLIP by capturing detailed features such as full asymmetry and symmetry in 1 axis. Furthermore, our method effectively highlights subtle variations, as illustrated in Figure \ref{fig:atypical_finetuned}, capturing a slight "light brown area" outside the atypical nevus region along with "symmetry in 1 axis" and other potentially significant features that the original Grad E-CLIP approach overlooked. Similarly, in Figure \ref{fig:melanoma_finetuned}, our method, MedGrad E-CLIP, excels in explaining "melanoma" and associated features like "fully symmetric" and "light brown color," even when features are absent or missing, thereby providing no explainability for those absent features.




\section{Conclusion}

This paper proposed MedGrad E-CLIP, an enhanced version of Grad E-CLIP that works well for medical data especially skin lesion dataset and which is able to capture subtle changes. The trained CLIP model on custom skin lesion dataset not only achieves improved accuracy but also generates more precise and relevant visualizations. We compared our proposed MedGrad E-CLIP with existing gradient-based methods, such as Grad-CAM and Grad E-CLIP, and showed that MedGrad E-CLIP provides superior visual explanations ensuring that it is able to capture fine-grained details by highlighting regions in the images more accurately and features that align well with their textual descriptions, thereby making the model's predictions more interpretable and reliable.


This work has certain limitations, particularly in the explainability of image-text pair relevance for cases where alignment between images and textual descriptions may lack clarity. Moreover, there are some cases where our proposed method excessively highlights features outside the intended regions, potentially leading to over-explanation.

In future work, we will focus on expanding the dataset and incorporating more detailed textual annotations for each skin lesion to enhance explainability and strengthen correlations between image and text pairs, thereby refining model alignment. Additionally, clinical trials will be conducted in collaboration with dermatologists to assess and validate the model’s applicability in real-world clinical settings, offering critical insights into its performance and reliability in practical diagnostic scenarios. To further validate our approach's robustness and reliability, we will perform quantitative assessments, such as insertion and deletion analysis, and statistical significance tests, to evaluate the effectiveness of the entropy-weighted attention mechanism and its ability to highlight diagnostically significant regions. Finally, efforts will also be directed toward addressing the susceptibility of gradient-based explanations to adversarial attacks, aiming to improve their stability and robustness.

\section{Acknowledgment}
We thank the referees for their valuable feedback and suggestions.

\clearpage
{\small
\bibliographystyle{ieee_fullname}
\bibliography{egbib}
}

\end{document}